\documentclass{article}

\usepackage{arxiv}

\usepackage[utf8]{inputenc} 
\usepackage[T1]{fontenc}    
\usepackage{hyperref}       
\usepackage{url}            
\usepackage{booktabs}       
\usepackage{amsfonts}       
\usepackage{nicefrac}       
\usepackage{microtype}      
\usepackage{amsmath}
\usepackage{xcolor}         
\usepackage{cleveref}       
\usepackage{lipsum}         
\usepackage{graphicx}
\usepackage{doi}

\usepackage{multirow}
\usepackage{listings}
\definecolor{codegreen}{rgb}{0,0.6,0}
\definecolor{codegray}{rgb}{0.5,0.5,0.5}
\definecolor{codepurple}{rgb}{0.58,0,0.82}
\definecolor{backcolour}{rgb}{0.95,0.95,0.92}
\lstdefinestyle{pythonstyle}{
  backgroundcolor=\color{backcolour}, commentstyle=\color{codegreen},
  keywordstyle=\color{magenta},
  numberstyle=\tiny\color{codegray},
  stringstyle=\color{codepurple},
  basicstyle=\ttfamily\footnotesize,
  breakatwhitespace=false,         
  breaklines=true,                 
  captionpos=b,                    
  keepspaces=true,                 
  numbers=left,                    
  numbersep=5pt,                  
  showspaces=false,                
  showstringspaces=false,
  showtabs=false,                  
  tabsize=2
}

\usepackage{tcolorbox}

\definecolor{lightred}{RGB}{255, 235, 235}
\definecolor{lightgreen}{RGB}{235, 255, 235}
\definecolor{lightblue}{RGB}{235, 245, 255}
\definecolor{bordergray}{RGB}{200, 200, 200}

\lstdefinelanguage{custom}{
    basicstyle=\ttfamily\small,
    keywordstyle=\bfseries,
    commentstyle=\itshape,
    morekeywords={Thought, Act, Obs, Finish},
    morestring=[b]",
}

\tcbuselibrary{listings,breakable}
\tcbset{
    colframe=bordergray,
    colback=white,
    boxrule=0.5pt,
    arc=2mm,
    left=2mm,
    right=2mm,
    top=2mm,
    bottom=2mm,
    breakable
}

\newtcolorbox{RAGbox}[1][]{colback=lightred, title=RAG-based, #1}
\newtcolorbox{Statementbox}[1][]{colback=lightblue, title=Statement-Only, #1}
\newtcolorbox{Agentbox}[1][]{colback=lightgreen}

\usepackage{caption}
\usepackage{subcaption}

\title{Understanding Inequality of LLM Fact-Checking over Geographic Regions with Agent and Retrieval models}

\newif\ifuniqueAffiliation

\ifuniqueAffiliation 
\author{%
}
\else
\usepackage{authblk}

\author[1]{Bruno Coelho}
\author[1]{Shujaat Mirza}
\author[1]{Yuyuan Cui}
\author[2]{Christina Pöpper}
\author[1]{Damon McCoy}
\affil[1]{New York University}
\affil[2]{New York University Abu Dhabi}
\affil[ ]{\textit{\{bruno.coelho, shujaat.mirza, yc1880,  christina.poepper, mccoy\}@nyu.edu}}
\fi

\renewcommand{\headeright}{}
\renewcommand{\undertitle}{}

\hypersetup{
pdftitle={A template for the arxiv style},
pdfsubject={q-bio.NC, q-bio.QM},
pdfauthor={David S.~Hippocampus, Elias D.~Striatum},
pdfkeywords={First keyword, Second keyword, More},
}

\begin{document}

\maketitle

\begin{abstract}

Fact-checking is a potentially useful application of Large Language Models (LLMs) to combat the growing dissemination of disinformation. However, the performance of LLMs varies across geographic regions.
In this paper, we evaluate the factual accuracy of open and private models across a diverse set of regions and scenarios.

Using a dataset containing 600 fact-checked statements balanced across six global regions we examine three experimental setups of fact-checking a statement: (1) when just the statement is available, (2) when an LLM-based agent with Wikipedia access is utilized, and (3) as a best case scenario when a Retrieval-Augmented Generation (RAG) system provided with the official fact check is employed. 
Our findings reveal that regardless of the scenario and LLM used, including GPT-4, Claude Sonnet, and LLaMA, statements from the Global North perform substantially better than those from the Global South. 
Furthermore, this gap is broadened for the more realistic case of a Wikipedia agent-based system, highlighting that overly general knowledge bases have a limited ability to address region-specific nuances.    
These results underscore the urgent need for better dataset balancing and robust retrieval strategies to enhance LLM fact-checking capabilities, particularly in geographically diverse contexts.
\end{abstract}


\section{Introduction}

The digital age has amplified our access to information, but in the same stride, it has magnified the challenge of mis-/disinformation~\cite{caramancion2020exploration}. The resulting wave of misleading content risks eroding the very foundations of our informed societies. Now, more than ever, as societies navigate through an intricate web of data-driven decisions, the integrity and trustworthiness of our information sources are under scrutiny~\cite{shu2020combating,grimes2021medical,bovet2019influence}.

Large Language Models (LLMs) have experienced widespread adoption due to their advanced capabilities in processing complex information~\cite{paul2023chatgpt}. Recognizing their potential impact, there has been an increasing focus on aligning these models with facts through Reinforcement Learning from Human Feedback (RLHF) to avoid harmful content generations~\cite{ouyang2022training}. As users increasingly place trust in LLMs with the responsibility of discerning fact from fiction~\cite{lee2021towards,guo2022survey,wahle2022testing,aggarwal2020classification,pelrine2021surprising}, ensuring their factuality and fairness becomes a central concern.

The impact of mis-/disinformation varies globally, influenced by factors like regional digital literacy and the strength of local fact-checking ecosystems. 
Regions with lower digital literacy are often more susceptible to misinformation due to the lack of robust mechanisms to critically assess and verify digital content~\cite{mirza2023tactics}. 
Such disparities highlight the need for examining the performance of LLMs across different regions. 
If these models are not finely attuned to regional variations, they risk exacerbating existing informational inequities. 
Our study, therefore, delves into the geographic factuality variations of LLMs, addressing an important aspect of how these technologies can either mitigate or amplify challenges in diverse information ecosystems.
It is crucial to understand these disparities as they can disproportionately affect underrepresented regions where reliable fact-checking resources may be scarce.

To evaluate these issues, we employ a balanced dataset comprising 600 fact-checked statements distributed evenly across six global regions. 
We observe notable performance disparities in LLMs based on geographic origin, uncovering a concerning bias where all of the LLMs that we studied more accurately fact-checked statements from the Global North (e.g., North America) than those from the Global South (e.g. Africa).
Such biases in computational systems can have far-reaching implications, potentially exacerbating existing inequities and misinforming certain populations. 

Next, we explore methods to improve the accuracy of LLMs when fact-checking statements. Ideally, we sought to discover methods that provide improvements and parity when fact-checking statements from different regions of the world. Our methods focus on providing the LLMs with access to external knowledge bases that might be useful for fact-checking. We explore multiple configurations, including the baseline statement-only setups, retrieval-augmented generation (RAG), and agent-based fact-checking scenarios. 
By examining these configurations, we aim to understand the strengths and limitations of LLMs under varying degrees of access to external information.

The different scenarios reveal a complex landscape. LLMs demonstrate high accuracy (exceeding 98\,\%) when provided with the correct document, highlighting their potential as effective tools in controlled environments. 
Unfortunately, their performance declines significantly in agent-based scenarios where models interact with a Wikipedia external knowledge-base. 
These declines are further pronounced for statements originating from the Global South, adding to the already existing gap between the regions.
Such results emphasize the urgent need for geographically balanced datasets and model training to address these inequities.

Our major contributions include:
\begin{enumerate}
    \item We highlight significant geographic disparities in fact-checking accuracy, with statements from the Global South consistently under-performing compared to those from the Global North. By quantitatively uncovering these discrepancies, our research highlights the need for culturally aware and geographically diverse model training and evaluation, which is vital for achieving equitable computational systems worldwide. This contribution is pivotal in guiding future LLM advancements towards global fairness, ensuring equitable technological benefits across diverse international contexts.

    \item We explore whether adding external knowledge bases improves accuracy and reduces inequality across different fact-checking scenarios, including statement-only, RAG, and agent-based setups,  offering nuanced insights into the strengths and limitations of current LLMs in the presence of external knowledge bases.

\end{enumerate}

\section{Related Work}

The recent advances in LLMs have been studied for many NLP tasks such as text summarization~\cite{yang_exploring_2023, wei_zero-shot_2023}, entity recognition~\cite{omar_chatgpt_2023, hu_zero-shot_2023}, and question and answering~\cite{katz_gpt-4_2023, openai_gpt-4_2023, huh_are_2023, qin2023chatgpt}. 
While impressive, these models are not immune to limitations and may raise their own problems. 
OpenAI has warned, ``GPT-4 `hallucinates' facts and makes reasoning errors'' to some extent~\cite{openai_gpt-4_2023}, Anthropic claims its Claude-series of models ``can occasionally produce responses that are incorrect or misleading''~\cite{anthropic_claude_nodate}, and Meta warns users on the LLaMA 3.3 model page that it ``may in some instances produce inaccurate, biased or other objectionable responses to user prompts''\cite{meta-llama-llama-70b-instruct}. 
Alongside the advancements in NLP, there is a growing concern over the impact of digital misinformation\cite{lewandowsky_beyond_2017}. The last few years have seen its proliferation into subjects such as climate change~\cite{biddlestone_climate_2022, van_der_linden_inoculating_2017}, vaccinations, and COVID-19~\cite{vidgen2021understanding, 10.1145/3543507.3583388}. 
The capabilities of LLMs raise important questions about their role as both a mitigator and generator of misinformation, either at the user level or as part of a larger system.

\subsection{Fact-Checking Capabilities of LLMs.}

Most of the work looking at freely available products has focused on ChatGPT and textual fact-checking, although newer multimodal models provide the chance to also fact-check images or videos~\cite{geng2024multimodallargelanguagemodels}.
ChatGPT has been used for examining both vaccination and cancer misconceptions~\cite{deiana2023artificial, johnson2023using}, both finding it provides generally accurate information.
Prior work~\cite{hoes2023using} has analyzed ChatGPT's performance for fact-checking using an open Politifact dataset. 
The researchers find that the source of the information can add up to 10 percentage points to the classification with blogs and campaigns being the best and worst categories with 77.7\% and 64.0\% accuracy, respectively.

Huang et al.~\cite{huang2023chatgpt} studied ChatGPT's text generation abilities for classifying and justifying the detection of Hate Speech. 
They find the model often makes use of an ``unclear'' label even when prompted to give a binary answer and these instances correlate with the more implicit/subtle examples. 
This tendency towards the lack of a binary answer is consistent with our own results.
ChatGPT has also been shown to be able to evaluate the credibility of news sources with ratings that correlate with human expert judgments, even in the face of non-English and satirical content~\cite{yang2023large}.
Fine-tuned models based on the earlier open-sourced GPT-2 have been shown to generate better corrective messages than those generated by humans~\cite{10.1145/3543507.3583388}. 

Overall these works provide an understanding of primarily chatbot interfaces, while we leverage direct access to their APIs to explore other scenarios which can use LLMs as a component of fact-checking. 

\subsection{Regional Disparities in LLMs}
Linguistic, geographical, and cultural biases in LLM models and the datasets they are trained on have been identified as a persistent issue.
Joshi et al.~\cite{joshi-etal-2020-state} explored the diversity of popular NLP datasets, finding a lack of regional diversity and arguing that this imbalance skews model performance.
Datasets such as XGLUE~\cite{liang-etal-2020-xglue} and AmericasNLI~\cite{ebrahimi-etal-2022-americasnli} tackle such disparities primarily by providing resources in non-Western languages in which these models can be trained or evaluated.

Outside dataset creation, prior work has looked at directly measuring geographical biases in LLM outputs.
When asked a ranking-style geographical question, LLMs exhibit biases across a range of both subjective and objective topics~\cite{manvi-geographically-biased}.
More specifically, the authors find LLMs are biased against locations with lower socioeconomic conditions (e.g., Africa versus North America or Europe) and propose custom metrics for capturing the degree of bias.
Other work has also looked at regional differences at a country-wide level, finding regional differences in Germany~\cite{kruspe-german-federal}.

In the realm of fact-checking, Min et al~\cite{shafayat2024multifact} propose Multi-FAct, a pipeline for measuring factuality of a
multilingual LLM and an extension of FActScore~\cite{min-etal-2023-factscore}.
In contrast to our work, the model is prompted to generate free-form text (e.g., the biography of national leaders) instead of being directly asked to fact-check, and the resulting text is then parsed for facts and evaluated against Wikipedia.
The authors find that GPT-4 performs better when queried in English, as well as when producing content related to North America and Europe, in line with our own findings.

\subsection{Fact-Checking NLP systems}

Besides the usage of LLM-powered chatbots, such models can also be used as part of a larger fact-checking system for classification, embedding, retrieval, or summarization of claims and documents.
Retrieval-Augmented Generation (RAG) frameworks enable models to dynamically query external databases to improve factual accuracy~\cite{lewis-retrieval-augmented-for-knowledge}.
In theory, this helps both with \emph{grounding} an answer with specific factual information to avoid hallucinations as well as allows a model to go beyond its training data cutoff~\cite{semnani-etal-2023-wikichat}.

Prior work has looked at the grounding phenomenon in the context of biomedical data~\cite{glockner-etal-2024-missci-dataset}, finding that RAG fact-checking systems often fail to find the right passage in a document and make LLMs more likely to accept a false claim as true~\cite{glockner2024grounding-findings}.

Approaches to make RAG systems better include a modular plug-and-play system with automated feedback~\cite{peng2023checkfactstryagain}, breaking a statement into multiple individual claims or questions and combining them either sequentially~\cite{momii-etal-2024-rag} or hierarchically~\cite{zhang-gao-2023-towards}, and adding multiple rounds of sequential question-answering~\cite{khaliq-etal-2024-ragar}.
In this paper, we use a ``best-case'' scenario where the retrieval part of a RAG system always returns the correct document to compare it with other approaches.

Another approach which has gained traction as a way to improve task-specific performance is using LLM-based agents for question-answering type tasks.
In this scenario, the agent is able to use ``tool calling''~\footnote{In this paper we use ``tool calling'' and ``function calling'' interchangeably.} to directly choose when and how to use complex functions that add functionality to the system.
Yao et al.~\cite{yao2023react} propose the ReAct methodology, which interweaves textual reasoning outputs (``As an LLM, I should ...'') with task-specific actions (``call function X with arguments Y'') for better long-term planning as well as increased interpretability. 
They evaluate their approach on a range of tasks, including the fact verification benchmark FEVER~\cite{Thorne18Fever}, with an agent capable of querying Wikipedia, finding that ReAct outperforms text-based action generation models.
We use a similar agent for our own experiments.

Quelle et al.~\cite{quelle-perils-promises} use a ReAct-like agent with google-search functionality on a custom multilingual dataset and the U.S.-focused PolitiFact dataset~\cite{vo-lee-2020-facts-politifact}. 
The authors find that GPT-4 outperforms GPT-3.5 and that performance is increased when querying the models in English, even when the original claim is in a different language that is supported by the model.
Also using primarily a dataset with statements from PolitiFact and a variation of FEVER~\cite{Thorne18Fever}, Tian et al. find adding RAG or search functionality increases performance of fact-checking. 
While these works have studied agent-based fact-checking, our focus is on the performance of models across statement from varied geographical regions.

\subsection{Other}
Other work~\cite{deverna2023artificial} has looked at the effectiveness of ChatGPT not just on performance but also on the belief and sharing intent of political U.S. news stories on social media style websites. 
When presented with a model-generated long-form textual explanation of a news headline, the authors find that while the model can accurately detect false content, it has a small or negative effects on sharing intent when compared to the control group, highlighting it's ineffectiveness as an intervention against misinformation.
The general effectiveness of warning labels has been questioned regardless of whether its human or AI-generated, as it may be inefficient~\cite{pantazi_power_2018,lee-warning-labels-covid} or have unexpected adverse effects~\cite{pennycook_implied_2020}, while other works indicate it may inoculate against false content~\cite{van_der_linden_inoculating_2017}.
The correction of misinformation has also been studied in connection to social ties~\cite{malhotra_facing_2022, malhotra_meaning_2022}, where technological approaches and fact-checking systems have been questioned regarding their usefulness~\cite{scott_i_2023}.

\section{Methodology}

\subsection{Dataset}

We expand on the original \emph{Global-Liar} dataset~\cite{mirza2024globalliar}.
This dataset focuses on six global regions:
 Africa, Asia-Pacific, Europe, Latin America, North America, and the Middle East with 100 statements collected from each region. 
Similarly to prior work using this dataset, we classify Africa, Latin America, and the Middle East as the Global South, while the Global North encompasses North America, Europe, and Asia-Pacific, although we recognize that the latter group comprises countries spanning both regions. 
In total, the dataset contains 600 examples of fact-checked statements from various new sources, equally balanced both across six geographical regions (Africa, Asia-Pacific, Europe, Latin America, North America, and the Middle East) and true and false statements.

Each example at its core a single statement, defined as a single sentence claim taken directly from an article, that summarizes and appears at the top half of the article.
Often, the statement is in the headline or a sentence in the very first lines of the article. 
The statements are pulled from the April 2017 to September 2023 date range. These dates are before the cutoff for all LLM models,~\footnote{October 2023, December 2023 and April 2024 for GPT-4o, LLaMA 3.3 and Claude 3.5 Sonnet respectively} which makes it possible that the models have previously been trained on the statements.

To achieve a balanced analysis, our dataset maintains an equal number of true and false statements. 
The False statements are sourced from AFP FactCheck~\footnote{\url{https://factcheck.afp.com/}}, a news agency that provides verification of a wide range of news articles.
True articles are obtained from news outlets from each region that are ranked highly on the International Media and Newspapers list~\footnote{\url{https://www.4imn.com/about/}}.
Regardless of factuality, statements are picked for their brevity and suitability for being addressed solely through textual content provided in the accompanying article. 
The original articles may mention images or videos but they are not required to be viewed to understand the factuality claim being made. 

After obtaining all statements, we crawl the article links with the intent of obtaining the full article text.
Our crawler then accesses a website in the browser and captures all readable text from a page through its HTML.
We then manually check the text extracted from all 600 articles and fix any mistakes, including stripping out unrelated text (e.g.,  headers, footers, comments, and ads) and manually copying text for the websites that block crawlers. 
We substitute 48 articles from the origithat were either completely inaccessible by link or had fewer than 300 characters.
When substituting, we maintain their respective region of origin and month of release for the new article.

\subsection{Models}

We focus our attention on publicly available and private models that can perform \emph{tool calling} as well as text generation.
We consider three private providers and their respective best-performing models, OpenAI and GPT-4o~\cite{openai-gpt4o-2024}, Anthropic and Claude Sonnet 3.5~\cite{anthropic-introducing-claude-35-2024}, and Google and Gemini 1.5 Pro~\cite{google-gemini-15-2024}.
Experiments showed the Gemini series to be subpar at function calling, often failing to use them so we exclude it from our models. 
We note that at the time of writing, a newer Gemini Flash 2~\footnote{\url{https://ai.google.dev/gemini-api/docs/models/gemini-v2}} became available that claims improved agentic capabilities.
For open-weight models, we focus on the LLaMA series 3~\cite{meta-llama-llama-70b-instruct}.

We use the latest model available for each API service, namely ``gpt-4o-2024-08-06'' (described as ``GPT-4o'' throughout the paper) and ``claude-3-5-sonnet-20241022'' (described as ``Claude 3.5'') from OpenAI and Anthropic, respectively.
For the LLaMA model, we use the LLaMA 3.3 70B Instruct version~\footnote{\url{https://huggingface.co/meta-llama/Llama-3.3-70B-Instruct}} (described as LLaMA 3.3) and hosted through Groq~\footnote{\url{https://groq.com/}}. 
We note all models are in the top 60 by rank on the Berkeley Function-Calling Leadearboard~\cite{berkeley-function-calling-leaderboard}, a benchmark of tool calling abilities of open and private models.

The LLaMA model is stated to run with 8-bit weights and FP16 arithmetic. 
All models are queried with the system message listed in Listing~\ref{listing:system-prompt}, a maximum user message size of 2000 characters (with any extra context truncated), a temperature setting of 0, and otherwise default settings.
The API queries were executed between October 2024 and January 2025. 

\begin{lstlisting}[
    caption={System prompt used for querying models. When running without tool-calling, the wikipedia functions line is removed.}, 
    label={listing:system-prompt},
    basicstyle= \ttfamily\small,  % Smaller font size
    breaklines=true,
    columns=fullflexible,
    frame=single,
    backgroundcolor=\color{gray!10},  % Lighter background color
    framexleftmargin=0.2em,  % Reduced margin
    framextopmargin=0.2em,   % Reduced margin
    framexbottommargin=0.2em  % Reduced margin
]
I need you to assume the role of a professional fact checker with an experience of 10+ years. You will fact-check a statement.

Task: For a given statement, delimited by triple backticks, rate it for truthfulness. 

Format the response as ``false'', ``true'', or ``unclear''.

You may use the `fetch_wikipedia_entity` and `fetch_wikipedia_entity_with_header` tool to help you acquire information.

You ***MUST*** only respond with one of these options and provide no context or explanations.
\end{lstlisting}

\subsection{Experiments}

We run three different scenarios for fact-checking: statement-only, RAG-based, and agent-based fact-checking.
These scenarios simulate varying levels of access to external information and functionality.
Modern products that provide user interfaces to LLMs, such as ChatGPT, now provide both the ability to upload documents and search the web in an agent-like behavior~\footnote{\url{https://help.openai.com/en/articles/8555545-file-uploads-faq} and \url{https://openai.com/index/introducing-chatgpt-search/} respectively}.
Our experiments, therefore, capture the functionality of fact-checking systems as well as possible user-driven workflows.

One form of structuring the output from a fact-checking model is to ask it to produce a ``true'' or ``false'' label. 
Early experimentation however showed trying to force such a behavior led to a wide range of unclear-like answers, formatted in different ways.
As such, in addition to the ``true'' and ``false'' labels, our prompt allows the language model to output an ``unclear'' verdict so it is more easily parsed.


\subsubsection{Statement-only Fact-Checking}

In this scenario, we provide a LLM only its system-level instructions and a user message with the statement to be fact-checked.
This evaluates the factuality of base models using only its internal knowledge, acts as a baseline for fact-checking systems, and captures how a user might interact with an LLM.
The leveraged user prompt is available in Listing~\ref{listing:user-prompt}.

\begin{lstlisting}[
    caption={User message prompt used for the statement-only and agent-based experiments.}, 
    label={listing:user-prompt},
    basicstyle= \ttfamily\small,  % Smaller font size
    breaklines=true,
    columns=fullflexible,
    frame=single,
    backgroundcolor=\color{gray!10},  % Lighter background color
    framexleftmargin=0.2em,  % Reduced margin
    framextopmargin=0.2em,   % Reduced margin
    framexbottommargin=0.2em  % Reduced margin
]
Now, let's get to task. Here is the statement: ```{statement}```
Please rate the statement as ``false'', ``true'', or ``unclear''.
\end{lstlisting}

\subsubsection{RAG-based Fact-Checking}
We simulate a RAG system where only a single document is retrieved, and it is always the correct news article relevant to the statement to be fact-checked.
In this configuration, the system's database is guaranteed to contain the required document, which is ranked as the most similar to the fact-checking query. 
We include the user prompt in Listing~\ref{listing:user-rag-prompt}.
By providing the LLM with an article known to contain the statement we are fact-checking, we obtain an upper-bound on an untuned RAG-based framework. 
Besides the high quality of the recommended article, the performance of the system is also limited by only considering statements that have already been reported on, failing to capture \emph{live} fact-checking claims that are critical in many real situations~\cite{glockner-etal-2022-missing}.

\begin{lstlisting}[
caption={User message prompt used for the RAG-based experiments.}, 
    label={listing:user-rag-prompt},
    basicstyle= \ttfamily\small,  % Smaller font size
    breaklines=true,
    columns=fullflexible,
    frame=single,
    backgroundcolor=\color{gray!10},  % Lighter background color
    framexleftmargin=0.2em,  % Reduced margin
    framextopmargin=0.2em,   % Reduced margin
    framexbottommargin=0.2em  % Reduced margin
]
Now, let's get to task. Here is the statement: ```{statement}```

To rate the statement, use the following related news articles: ```{full_article}``` 

Please rate the statement as ``false'', ``true'', or ``unclear''.
\end{lstlisting}

\subsubsection{Agent-based Fact-Checking}
In the agent-based scenario, we create a ReAct-like~\cite{yao2023react} agent, capable of querying Wikipedia, using the LangGraph library~\footnote{\url{https://langchain-ai.github.io/langgraph/}}.
The agent's main component is an LLM capable of function calling, instantiated with a system message and bound to two predefined tools so that the agent will always send structured function schemas on subsequent LLM calls. 
Advances in LLM training mean that we differ from the original ReAct framework by having more complex functionalities built directly into the model.
Namely, we use native tool calling, interact through specific types of messages (system, user, and AI), don't require all inputs to be a single string, and don't have a dedicated reasoning step since models are now often trained specifically for tool usage.

We create two Python functions the model can use:
\begin{itemize}
    \item \emph{``fetch\_wikipedia\_entity'':} Queries Wikipedia for articles regarding a specific entity. 
    If multiple pages share an entity name, the top 10, as ranked by Wikipedia, are returned. If no pages are found, the empty search is returned.
    When a match is found, the article's textual content is parsed from the page's HTML.  If the page exceeds the character limit, the agent is instructed to use the function below. 
    \item \emph{``fetch\_wikipedia\_entity\_with\_header'':} Returns the textual content under a specified header.  If the header doesn't appear in the article, the function responds with an error message indicating the valid headers.
\end{itemize}

For both functions, we implement a local cache system to avoid querying Wikipedia multiple times for the same page. Due to this implementation, the version of the Wikipedia page the model saw can be up to 3 months older than the date of execution.
The agent can query use the functions multiple times and at any time may stop to provide a final answer.
We limit the models to a maximum of 15 function call steps. In case the model hits that limit, we return the message, \emph{``You have reached the maximum number of model calls. Please rate the statement as `false', `true', or `unclear' and do not perform any more function calls.''}.
The first user message is the same as the statement-only scenario.
An example trajectory~\footnote{We use ``trajectory'' to refer to the sequence of observations, actions, and decisions made by the agent during its operation.}, shortened for brevity, is shown in Listing~\ref{listing:sample-short-trajectory} with a complete example  in Appendix~\ref{app:agent-example} and the full function signature and doc-string for the tools in Appendix~\ref{app:tool-calling}.

\begin{lstlisting}[
caption={Example trajectory for the agent-based scenario with GPT-4o as the model. The statement being fact-checked is ``Controversial quotes made by Russian President Vladimir Putin criticising the lifestyle of  `rich Africans' '' from \protect\hyperlink{https://factcheck.afp.com/no-evidence-behind-claims-vladimir-putin-made-statement-criticising-rich-africans}{AFP Fact-Check}. Edited content beween square brackets.}, 
    escapeinside={(*@}{@*)}, % Define escape markers for LaTeX commands
    label={listing:sample-short-trajectory},
    basicstyle= \ttfamily\small,  % Smaller font size
    breaklines=true,
    columns=fullflexible,
    frame=single,
    backgroundcolor=\color{gray!10},  % Lighter background color
    framexleftmargin=0.2em,  % Reduced margin
    framextopmargin=0.2em,   % Reduced margin
    framexbottommargin=0.2em  % Reduced margin
]
(*@\textbf{AI Message:}@*) Tool Calls: fetch_wikipedia_entity_with_header(entity=``Vladimir Putin'', header=``Controversies'')
(*@\textbf{Observation:}@*) Could not find Controversies in the Wikipedia page. Valid headers are: [LIST OF HEADERS]

(*@\textbf{AI Message:}@*) Tool Calls: fetch_wikipedia_entity_with_header(entity=``Vladimir Putin'', header=``Racism and xenophobia'')
(*@\textbf{Observation:}@*) Could not find Racism and xenophia in the Wikipedia page. Valid headers are: [LIST OF HEADERS]

(*@\textbf{AI Message:}@*) Tool Calls: fetch_wikipedia_entity_with_header(entity=``Vladimir Putin'', header=``Public image'')
(*@\textbf{Observation:}@*) The director of the Levada Center stated in 2015 that ... [(*@\hyperlink{https://en.wikipedia.org/wiki/Vladimir_Putin\#Public_image}{CONTENT FROM THE WIKIPEDIA PAGE}@*)]

(*@\textbf{AI Message:}@*) false
\end{lstlisting}


\section{Results}

For the examples where a model hits the limit of API calls and still doesn't return a valid classification, we set their answers to ``unclear''. 
This happened 12 times for GPT-4o. Sonnet 3.5 never reached the API limit and Llama 3.3 exceeded it 6 times, but still provided a decision.

\subsection{Regional Analysis}

The disparity between the Global North and Global South is evident in Table~\ref{tab:alt}. 
With the exception of Europe being one percentage point below the Middle East for the RAG-based Sonnet 3.5 experiment, performance from the three regions in the Global South always make up the three worst performing regions.
On average, the Global North has 29.5\,\% higher accuracy than the Global South.

When breaking down by specific regions, North America consistently achieves the highest accuracy rates, except for the statement-only GPT-4o scenario, which is one percentage point below Europe.

Our analysis proceeds with the fitting of a logistic regression model to evaluate the performance of the different models across regional differences and fact-checking scenarios. 
We consider instances where the model output was ``unclear'' as incorrect so that a binary label is produced for each statement and experiment.
Our logistic regression model delineates individual regions as standalone categories, providing insights into the region-specific performance of each experiment. 
Details of this model are presented in Table~\ref{tab:log-reg-results}, with Appendix~\ref{app:log-reg} providing further information.


Logistic regression analyses reveal statistically significant geographic disparities in model performances. In individual region assessments, 
the Global South exhibits a marked underperformance when contrasted with the Global North. 
Specifically, North America performs better than the reference, Asia-Pacific, whereas Africa, Latin America, and the Middle East lag significantly behind, as indicated by their negative coefficients and statistical significance (p $\leq$ 0.01, with Europe performing better but not statistically significant so than Asia-Pacific). 
The substantial z-scores associated with these coefficients confirm the robustness of these disparities.

In summary, there is a clear indication of geographic disparities in model performance, with the Global North, particularly North America, receiving the most benefit from model accuracies.
Regions such as Africa and the Middle East are at a disadvantage, with much lower accuracy, pointing to the need for more representative training datasets.
Additionally, the performance fluctuations over time for all regions suggest that model updates may not consistently benefit all areas equally, which is an important consideration for the ongoing development and deployment of LLMs.

\begin{table}[tb]
  \caption{Accuracy results across fact-checking scenario, regions, and models. Minimum accuracy results per model highlighted in bold.}
  \label{tab:alt}
  \centering
  \begin{center}
    \addtolength{\leftskip} {-2cm} 
    \addtolength{\rightskip}{-2cm}
\begin{tabular}{lccc|ccc|ccr}
\toprule
\multirow{2}{*}{\textbf{Region}} & \multicolumn{3}{c}{\textbf{Statement-only}} & \multicolumn{3}{c}{\textbf{Agent-based}} & \multicolumn{3}{c}{\textbf{RAG-based}} \\
\cmidrule(lr){2-4} \cmidrule(lr){5-7}
\cmidrule(lr){8-10}
& GPT-4o & Sonnet 3.5 & LLaMA 3.3 & GPT-4o & Sonnet 3.5 & LLaMA 3.3 & GPT-4o & Sonnet 3.5 & LLaMA 3.3 \\
\midrule
Africa & \textbf{43.0\,\%} & \textbf{66.0\,\%} & \textbf{67.0\,\%} & \textbf{19.0\,\%} & \textbf{24.0\,\%} & \textbf{58.0\,\%} & 98.0\,\% & \textbf{98.0\,\%} & 99.0\,\% \\
Asia-Pacific & 73.0\,\% & 87.0\,\% & 84.0\,\% & 42.0\,\% & 46.0\,\% & 85.0\,\% & 100\,\% & 99.0\,\% & 100\,\% \\
Europe & 76.0\,\% & 92.0\,\% & 87.0\,\% & 33.0\,\% & 69.0\,\% & 87.0\,\% & 99.0\,\% & \textbf{98.0\,\%} & 99.0\,\% \\
Latin America & 60.0\,\% & 82.0\,\% & 83.0\,\% & 25.0\,\% & 40.0\,\% & 82.0\,\% & 97.0\,\% & \textbf{98.0\,\%} & \textbf{97.0\,\%} \\
Middle East & 63.0\,\% & 85.0\,\% & 76.0\,\% & 26.0\,\% & 38.0\,\% & 76.0\,\% & 98.0\,\% & 99.0\,\% & 99.0\,\% \\
North America & 75.0\,\% & 97.0\,\% & 96.0\,\% & 52.0\,\% & 74.0\,\% & 93.0\,\% & 100\,\% & 100\,\% & 100\,\% \\
\midrule
Global North & 74.7\,\% & 92.0\,\% & 89.0\,\% & 42.3\,\% & 63.0\,\% & 88.3\,\% & 99.7\,\% & 99.0\,\% & 99.7\,\% \\
Global South & \textbf{55.3\,\%} & \textbf{77.7\,\%} & \textbf{75.3\,\%} & \textbf{23.3\,\%} & \textbf{34.0\,\%} & \textbf{72.0\,\%} & \textbf{97.7\,\%} & \textbf{98.3\,\%} & \textbf{98.3\,\%} \\
\midrule
Total & 65.0\,\% & 84.8\,\% & 82.2\,\% & 32.8\,\% & 48.5\,\% & 80.2\,\% & 98.7\,\% & 98.7\,\% & 99.0\,\% \\
\bottomrule
\end{tabular}
\end{center}
\end{table}

\begin{table}[htbp]
\centering
  \caption{Logistic regression coefficients, standard errors, and p-values. The reference values are Sonnet 3.5, Asia-Pacific and statement-only for the model, region, and scenario variables respectively.}
  \label{tab:log-reg-results}
\begin{center}
\begin{tabular}{lcccccr}
\toprule
  \textbf{Variables}  & \textbf{Coefficient} & \textbf{Standard Error} & \textbf{z} & \textbf{P$> |$z$|$} & \textbf{[0.025} & \textbf{0.975]}  \\
\midrule
\textbf{Intercept}                                                         &       1.6112  &        0.120     &    13.445  &         0.000        &        1.376    &        1.846     \\
\textbf{Model = GPT-4o}             &      -0.8293  &        0.090     &    -9.229  &         0.000        &       -1.005    &       -0.653     \\
\textbf{Model = LLaMA 3.3}          &       0.8592  &        0.101     &     8.498  &         0.000        &        0.661    &        1.057     \\
\textbf{Region = Africa}          &      -1.1540  &        0.130     &    -8.881  &         0.000        &       -1.409    &       -0.899     \\
\textbf{Region = Europe}          &       0.2250  &        0.137     &     1.641  &         0.101        &       -0.044    &        0.494     \\
\textbf{Region = Latin America}   &      -0.4425  &        0.131     &    -3.377  &         0.001        &       -0.699    &       -0.186     \\
\textbf{Region = Middle East}     &      -0.4748  &        0.131     &    -3.628  &         0.000        &       -0.731    &       -0.218     \\
\textbf{Region = North America}   &       0.7335  &        0.146     &     5.026  &         0.000        &        0.447    &        1.020     \\
\textbf{Scenario = Agent-based} &      -1.2537  &        0.081     &   -15.532  &         0.000        &       -1.412    &       -1.096     \\
\textbf{Scenario = RAG-based}   &       3.3453  &        0.225     &    14.891  &         0.000        &        2.905    &        3.786     \\
\bottomrule
\end{tabular}
\end{center}
\end{table}

\subsection{Model and fact-checking scenarios}

We next turn our attention to differences in between models and fact-checking scenarios, with two key insights.

First, when provided directly with the correct document, all evaluated models achieve a very high accuracy, ranging from 97\,\% to 100\,\%. 
This indicates LLMs may be particularly useful in scenarios where a reliable dataset of pre-fact-checked content is available.

In contrast, the performance can deteriorate significantly in the agent-based scenario with Wikipedia access, as seen by the statistically significant negative coefficient for the agent-based statements in the logistic regression.
More specifically, both GPT-4o and Sonnet 3.5 perform worse than randomly guessing \emph{true} or \emph{false}, at 32.8\,\% and 48.5\,\% respectively.
The best performing model, LLaMA 3.3 still gets approximately one in every five claim wrong, with twice as many for claims from the African region.


\begin{table}[hbtp]
  \caption{Model Analysis Summary for multiple scenarios. The rows ``True'' and ``False'' represent the correct label that should be given, while the columns represent the models predictions.}
  \label{tab:error-analysis}

\begin{subtable}[c]{\textwidth}
\centering
\caption{RAG-based scenario.}
\label{tab:error-analysis-a-statement}
\begin{tabular}{lccccccccr}
\toprule
\multirow{2}{*}{\textbf{Actual}} & 
\multicolumn{9}{c}{\textbf{Predicted}} \\ 
\cmidrule(lr){2-10}
 & \multicolumn{3}{c}{\textbf{GPT-4o}} & \multicolumn{3}{c}{\textbf{Sonnet 3.5}} & \multicolumn{3}{c}{\textbf{LLaMA 3.3}} \\
 & \textbf{True} & \textbf{False} & \textbf{Unclear} & \textbf{True} & \textbf{False} & \textbf{Unclear} & \textbf{True} & \textbf{False} & \textbf{Unclear} \\
\midrule
\textbf{True} & 297 & 1 & 2 & 295 & 1 & 4 & 299 & 1 & 0 \\
\textbf{False} & 4 & 295 & 1 & 2 & 297 & 1 & 4 & 295 & 1 \\
\midrule
\textbf{Total} & 301 & 296 & 3 & 297 & 298 & 5 & 303 & 296 & 1 \\
\bottomrule
\end{tabular}

\end{subtable}

\vspace*{0.3 cm}
  
\begin{subtable}[c]{\textwidth}
\centering
\caption{Agent-based scenario.}
\label{tab:error-analysis-b-agent}
\begin{tabular}{lccccccccr}
\toprule
\multirow{2}{*}{\textbf{Actual}} & 
\multicolumn{9}{c}{\textbf{Predicted}} \\ 
\cmidrule(lr){2-10}
 & \multicolumn{3}{c}{\textbf{GPT-4o}} & \multicolumn{3}{c}{\textbf{Sonnet 3.5}} & \multicolumn{3}{c}{\textbf{LLaMA 3.3}} \\
 \cmidrule(lr){2-4} \cmidrule(lr){5-7} \cmidrule(lr){8-10}
 & \textbf{True} & \textbf{False} & \textbf{Unclear} & \textbf{True} & \textbf{False} & \textbf{Unclear} & \textbf{True} & \textbf{False} & \textbf{Unclear} \\
\midrule
\textbf{True} & 84 & 3 & 213 & 111 & 7 & 182 & 256 & 26 & 18 \\
\textbf{False} & 4 & 113 & 183 & 17 & 180 & 103 & 69 & 225 & 6 \\
\midrule
\textbf{Total} & 88 & 116 & 396 & 128 & 187 & 285 & 325 & 251 & 24 \\
\bottomrule
\end{tabular}
\end{subtable}

\vspace*{0.3 cm}

\begin{subtable}[c]{\textwidth}
\centering
\caption{Statement-only scenario.}
\label{tab:error-analysis-c-rag}
\begin{tabular}{lccccccccr}
\toprule
\multirow{2}{*}{\textbf{Actual}} & 
\multicolumn{9}{c}{\textbf{Predicted}} \\ 
\cmidrule(lr){2-10}
 & \multicolumn{3}{c}{\textbf{GPT-4o}} & \multicolumn{3}{c}{\textbf{Sonnet 3.5}} & \multicolumn{3}{c}{\textbf{LLaMA 3.3}} \\
 & \textbf{True} & \textbf{False} & \textbf{Unclear} & \textbf{True} & \textbf{False} & \textbf{Unclear} & \textbf{True} & \textbf{False} & \textbf{Unclear} \\
\midrule
\textbf{True} & 220 & 5 & 75 & 264 & 6 & 30 & 286 & 11 & 3 \\
\textbf{False} & 21 & 170 & 109 & 36 & 245 & 19 & 89 & 207 & 4 \\
\midrule
\textbf{Total} & 241 & 175 & 184 & 300 & 251 & 49 & 375 & 218 & 7 \\
\bottomrule
\end{tabular}
\end{subtable}

\end{table}

Investigation of the error cases indicates that private models are likelier to answer \emph{unclear}.
The findings in Table~\ref{tab:error-analysis-b-agent} highlight the over 250 statements (out of 600) that were labeled as unclear in the agent-based scenario by both GPT-4o and Sonnet 3.5.
The former classifies 396 statements, almost two in every three, of all statements as unclear, up from 184 in the statement-only scenario.
While claiming a False claim is Unclear is a lesser error than claiming it True (and likewise for a True claim), prior work has found LLM fact-checking systems that increase belief in dubious headlines lead to a higher intent to share false claims and an increased belief of dubious headlines~\cite{doi:10.1073/pnas.2322823121-DeVerna-headline}. They also find that the ChatGPT user-interface, likely powered by a model very similar to GPT-4o, answered Unclear for 90\% of true statements and 10\% for false statements.

Although all models also lower the number of false positives (i.e. true examples that are predicted as false) by one to twenty statements, this is offset by the large amount of claims that the models are unsure about. 
LLaMA 3.3 was the least impacted model, still having a decrease in overall accuracy when compared to the statement-only scenario, but of only 2\% instead of the 30+ for the other models. 
This is mainly due to the much smaller increase in unclear claims.

To summarize, these results indicate that this more realistic scenario — where the database does not necessarily contain the exact document matching the claim and if it does, it must be found and extracted — introduces challenges that hinder the models’ fact-checking capabilities. 
Furthermore, using a general knowledge base seems to not only result in decreases in performance, but also comes with the cost of exacerbating regional differences as well as the risk of decreasing trust in LLM-based fact-checking systems due to an increased hesitancy in the model to make a definitive classification. 


\section{Discussion}

The findings in this study reveal both the strengths and weaknesses of current LLMs-based systems for fact-checking.

\subsection{Lessons Learned}

In a controlled environment where LLMs are provided the correct document, our analyses show that they can achieve impressive accuracies at or above 97\,\%, indicating their potential when paired  with curated databases of pre-verified content. 
While LLaMA 3.3 outperformed other models, it only did so by 0.3 percentage points, highlighting that in such scenarios the decision of which model to use may have little impact.

However, in the more realistic scenario where the model doesn't have a direct access to the answer, we observe large regional differences.
The underrepresentation of the Global South, across almost all scenarios and models, 
points to a significant bias in data sources or model performance.
We believe enriching future models with training data from these underrepresented areas is not only a matter of fairness but also a necessity for building globally competent fact-checking tools. 
Further research could explore the impact of more diverse training data on downstream tasks such as fact-checking.

In an agent-based scenario with access to Wikipedia, we observe that the added functionality does not lead to an improvement in performance.
The added complexity of querying and interpreting external sources often resulted in significant drops in accuracy, with GPT-4o and Sonnet 3.5 under-performing to the point of yielding results worse than random guessing.
For LLaMA 3.3, the region of Africa, already the one with the lowest performance, saw the biggest decline in accuracy.
This indicates that adding agent-capabilities with a general knowledge-base is not a viable solution to solving regional differences in these models.
Future work could analyze 


\subsection{Limitations}

We focus on the binary fact-checking problem and do not investigate how systems may use LLMs to combat misinformation through model-generated explanations or human interactions. 
While misinformation may come in many different formats, we focus on evaluating only textual claims. Furthermore, we focus on the quality of the generated labels and not how a user might perceive them. 
Prior research has shown that trust in AI systems depends on a wide range of factors including meta-information, system design, and personal factors~\cite{deverna2023artificial}. 
We also note that these models have the capability to generate misinformation, reducing trust and potentially being more harmful than good.

\section{Conclusion}


This study's examination of open and private models for fact-checking reveals significant disparities in factual accuracy and biases, particularly disadvantaging the Global South. 
The introduction of the 'Global-Liar' dataset underscores the need for geographically diverse data in AI training. 
The observed bias in favor of the Global North underscores a significant concern: existing AI models may perpetuate informational inequities, making the inclusion of underrepresented regions in AI training a priority. 

The agent-based scenarios raise important questions about whether relying on external knowledge bases, such as Wikipedia, introduces more challenges than benefits.
Our results shows that performance declines in agent-based scenarios that rely on such knowledge bases due to a tendency of models to return an unclear verdict.  

Overall, these insights stress the importance of inclusive and representative training for LLMs to ensure global fairness. 
As AI continues to influence information dissemination, this research advocates for a concerted effort toward developing AI systems that are equitable and effective across diverse global communities.



\bibliographystyle{ACM-Reference-Format}
\bibliography{references}

\appendix

\section{Research Methods}

\subsection{Tool Calling Functions}\label{app:tool-calling}

Listing~\ref{listing:fetch-entity-func} and Listing~\ref{listing:fetch-entity-with-headers-func} show the header and docstrings for the two functions used to interact with Wikipedia.

\lstset{style=pythonstyle}
\begin{lstlisting}[language=Python, caption={Python docstring for the ``fetch\_wikipedia\_entity'' function.},
label={listing:fetch-entity-func}
]
def fetch_wikipedia_entity(entity: str) -> str:
    """Search for a given entity on Wikipedia.
    Examples Usage: "fetch_wikipedia_entity('Python')" will return the content of the Python Wikipedia page.
    Use the `fetch_wikipedia_entity_with_header` tool to get specific information.

    Args:
        entity: The entity to search for.
    """
\end{lstlisting}
\begin{lstlisting}[language=Python, caption={Python docstring for the ``fetch\_wikipedia\_entity\_with\_header'' function.},
label={listing:fetch-entity-with-headers-func}
]
def fetch_wikipedia_entity_with_header(entity: str, header: str) -> str:
    """Get the content of a specific header from a Wikipedia page.
    This function should be used after `fetch_wikipedia_entity`. Examples Usage: "fetch_wikipedia_entity_with_header('Python', 'History')" will return the History section of the Python Wikipedia page.

    Args:
        entity: The entity to search for.
        header: The header to extract the content from.

    Returns:
        str: The content of the header or an error message.
    """
\end{lstlisting}

\subsection{Logistic Regression}\label{app:log-reg}

Table~\ref{tab:log-reg-details} and Table~\ref{tab:log-reg-results} contain further details on the logistic model used. Given the binary ``correct'' label, the full model is described by $correct  \sim C(model, Treatment(reference=`Sonnet 3.5')) + C(region, Treatment(reference=`Asia-Pacific')) + C(scenario, Treatment(reference=`Statement-Only')) + 1$.

\begin{table}[htbp]
  \caption{Logit regression model details.}
  \label{tab:log-reg-details}
\begin{center}
\begin{tabular}{lclc}
\toprule
\textbf{Dep. Variable:}   &     correct      & \textbf{  No. Observations:  } &    5400   \\
\textbf{Model:}           &      Logit       & \textbf{  Df Residuals:      } &    5390   \\
\textbf{Method:}          &       MLE        & \textbf{  Df Model:          } &       9   \\
\textbf{  Pseudo R-squ.:     }  & 0.2991 & \textbf{  Log-Likelihood:    } &  -2057.0  \\
\textbf{converged:}       &       True       & \textbf{  LL-Null:           } &  -2934.9  \\
\textbf{Covariance Type:} &    nonrobust     & \textbf{  LLR p-value:       } &   0.000   \\
\bottomrule
\end{tabular}
\end{center}
\end{table}

\subsection{Sample Executions}\label{app:agent-example}

In this section we list a sample trajectory for the agent-based scenario.
We skip the starting system and user message for brevity.
The agent may perform multiple function-calls in a single action but we edit the output to be grouped by Wikipedia page for easier comparison. 
The example is executed with a LLaMA-based agent.


\noindent\textbf{Statement:} ``British Prime Minister Boris Johnson and US Secretary of State Mike Pompeo have expressed support for the secession of Biafra -- a collection of states from southeast Nigeria -- from the rest of the country.'' A false statement obtained from \href{https://factcheck.afp.com/false-quotes-claim-boris-johnson-and-mike-pompeo-express-support-nigerian-separatists}{AFP FactCheck} in the Europe region.
%

\begin{Agentbox}
\lstinline[breaklines=true,breakindent=5mm]| |
\textbf{AI Message:} Tool Calls: fetch\_wikipedia\_entity(entity=``Boris Johnson'')

\textbf{Observation:} The page is too long. Use the `fetch\_wikipedia\_entity\_with\_header` tool to get specific information. Headers: ['Contents', 'Early life and education', 'Childhood', 'Eton and Oxford: 1977–1987', 'Early career', 'The Times and The Daily Telegraph: 1987–1994', 'Political columnist: 1994–1999', 'The Spectator and MP for Henley: 1999–2008', 'Becoming an MP', 'Second term', 'Mayor of London (2008–2016)', 'Mayoral election: 2007–2008', 'First term: 2008–2012', 'Policies', 'Relations with the police, finance, and the media', 'Re-election campaign', 'Second term: 2012–2016', 'Return to Parliament', 'Brexit campaign: 2015–2016', 'Foreign Secretary: 2016–2018', 'Return to the backbenches: 2018–2019', 'Journalism', '2019 Conservative Party leadership election', 'Premiership (2019–2022)', 'First term (July–December 2019)', 'Brexit policy', 'First Cabinet', 'Loss of working majority', '2019 general election', 'Second term (December 2019 – September 2022)', 'Second Cabinet', 'COVID-19 pandemic', 'Immigration', 'Legislative agenda', '2021 Downing Street refurbishment controversy', 'Owen Paterson controversy', 'Partygate scandal', 'Starmer slur controversy', 'Vote of confidence', 'June 2022 by-elections', 'Pincher scandal', 'Mass resignations', 'Announcement of resignation', 'Environmental policies', 'Foreign policy', 'Hong Kong and China', 'Afghanistan', 'UK–EU trade negotiation', 'Russia and Ukraine', 'Post-premiership (2022–present)', 'Political positions and ideology', 'Environment', 'Immigration and the European Union', 'Unionism and devolution', 'Public image', 'Personal life', 'Religion', 'Relationships', 'Family and ancestors', 'Honours', 'Works', 'Notes', 'References', 'Footnotes', 'Sources', 'Further reading', 'External links']

\textbf{AI Message:} Tool Calls: fetch\_wikipedia\_entity\_with\_header(entity=``Boris Johnson'', header=``Foreign policy'')

\textbf{Observation:} Johnson supported the European Union–Mercosur Free Trade Agreement,[571] which would form one of the world's largest free trade areas.[572] Johnson's government placed importance on the "Special Relationship" with the United States.[573][574] In 2022, his government introduced an asylum deal whereby people entering the UK illegally would be sent to Rwanda.[575]
Johnson said in July 2019 that his government would be very "pro-China" in an interview with the Hong Kong broadcaster Phoenix TV. He voiced support for Chinese president Xi Jinping's infrastructure investment effort, the Belt and Road Initiative, and promised to keep the United Kingdom "the most open economy in Europe" for Chinese investment.[576]
In June 2020 Johnson announced that if China were to continue pursuing the Hong Kong national security law, the UK would offer 350,000 Hong Kong residents who are British National (Overseas) passport holders, and 2.6 million other eligible individuals, the chance to move to the UK.[577] China accused the UK of interfering in its internal affairs.[578]
Johnson declined to describe the Chinese government's treatment of the Uyghur people as "genocide", despite use of the term by the United States.[579] Johnson's government argued that genocide should be decided by the International Criminal Court.[580] Nevertheless, he called what is happening to the Uyghurs in Xinjiang as "utterly abhorrent".[579]
The UK joined the AUKUS defence pact with the United States and Australia in September 2021. The pact was denounced by China[581] and caused a French backlash, as it usurped existing plans for Australia to procure French submarines.[582]
On 8 July 2021 the day after saying he was "apprehensive" about the future of Afghanistan following what was then the impending withdrawal of US troops, while announcing the near completion of British troop withdrawal from Afghanistan, Johnson expressed the view that there was "no military path to victory for the Taliban".[583][584] Following [MAXIMUM LENGTH]

\textbf{AI Message:} Tool Calls: fetch\_wikipedia\_entity(entity=``Mike Pompeo'')

\textbf{Observation:} The page is too long. Use the `fetch\_wikipedia\_entity\_with\_header` tool to get specific information. Headers: ['Contents', 'Early life and education', 'Early career', 'U.S. House of Representatives (2011–2017)', 'Elections', 'Tenure in Congress', 'CIA Director (2017–2018)', 'Secretary of State (2018–2021)', 'Nomination and confirmation', 'Tenure as secretary of state', 'Madison dinners', 'Threatening of the International Criminal Court', 'Inspector general investigations', 'Impeachment inquiry against Donald Trump', 'COVID-19 pandemic', '2019 emergency arms sale', 'Republican National Convention speech', 'Afghanistan and the Taliban', 'Final days in office', 'Post-Trump administration (2021–present)', 'Political positions', 'Foreign policy', 'China', 'Taiwan', 'Russia', 'Energy and environment', 'Health care', 'Social issues', 'Economy', 'International Criminal Court', 'Personal life', 'Foreign honors', 'See also', 'References', 'External links']

\textbf{AI Message:} Tool Calls: fetch\_wikipedia\_entity\_with\_header(entity=``Mike Pompeo'', header=``Foreign policy'')

\textbf{Observation:} In 2013, Pompeo supported the surveillance programs of the National Security Agency, referring to the agency's efforts as "good and important work".[194] In 2016 Pompeo stated, "Congress should pass a law re-establishing collection of all metadata, and combining it with publicly available financial and lifestyle information into a comprehensive, searchable database. Legal and bureaucratic impediments to surveillance should be removed. That includes Presidential Policy Directive-28, which bestows privacy rights on foreigners and imposes burdensome requirements to justify data collection."[195] In March 2017, WikiLeaks began publishing a series of documents known as Vault 7, detailing the CIA's electronic surveillance and cyber warfare activities and capabilities. In an April 2017 speech addressing the Center for Strategic and International Studies, Pompeo called WikiLeaks "a non-state hostile intelligence service" and described Assange as a "narcissist" and "a fraud—a coward hiding behind a screen".[196][197] In 2022 a lawsuit was filed against Pompeo and other defendants alleging that they have illegally placed Assange and his guests under surveillance.[198][199]
In a 2013 speech on the House floor, Pompeo said Muslim leaders who fail to denounce acts of terrorism done in the name of Islam are "potentially complicit" in the attacks.[200] The Council on American–Islamic Relations called on him to revise his remarks, calling them "false and irresponsible".[201] In 2016, ACT for America gave Pompeo a "national security eagle award" for his comments on Islam.[202] Pompeo has been a frequent guest on anti-Muslim activist Frank Gaffney's radio show for the Center for Security Policy.[202] As a congressman, he cosponsored legislation to add the Muslim Brotherhood to the United States State Department list of Foreign Terrorist Organizations.[203][204]
Pompeo opposed closing the Guantanamo Bay detention camp.[205] After a 2013 visit to the prison, he said, of the prisoners [MAXIMUM LENGTH]

\textbf{AI Message:} Tool Calls: fetch\_wikipedia\_entity(entity=``Biafra'')

\textbf{Observation:} The page is too long. Use the `fetch\_wikipedia\_entity\_with\_header' tool to get specific information. Headers: ['Contents', 'History', 'Under independent Nigeria', 'Nigerian civil war', 'Geography', 'Languages', 'Politics', 'International recognition', 'Economy', 'Military', 'Legacy', 'Post-war events and Biafran nationalism', 'Indigenous People of Biafra (IPOB)', 'Biafra Republic Government in Exile (BRGIE)', 'See also', 'Notes', 'References', 'Further reading', 'External links']

\textbf{AI Message:} Tool Calls: fetch\_wikipedia\_entity\_with\_header(entity=``Biafra'', header=``Post-war events and Biafran nationalism'')

\textbf{Observation:} The Movement for the Actualization of the Sovereign State of Biafra (MASSOB) emerged in 1999 as a nonviolent Biafran nationalist group, associated with Igbo nationalism. The group enacted a "re-launch" of Biafra in Aba, the commercial centre of Abia State and a major commercial centre on Igbo land.[67] MASSOB says it is a peaceful group and advertises a 25-stage plan to achieve its goal peacefully.[68] It has two arms of government, the Biafra Government in Exile and the Biafra Shadow Government.[69] MASSOB accuses Nigeria of marginalising Biafran people.[70] Since August 1999, protests have erupted in cities across Nigeria's south-east. Though peaceful, the protesters have been routinely attacked by the Nigerian police and army, with large numbers of people reportedly killed. Many others have been injured and/or arrested.[71]
On 29 May 2000, the Lagos Guardian newspaper reported that the former president Olusegun Obasanjo commuted to retirement the dismissal of all military persons, soldiers and officers, who fought for the breakaway Republic of Biafra during Nigeria's 1967–1970 civil war. In a national broadcast, he said the decision was based on the belief that "justice must at all times be tempered with mercy".[72]
In July 2006, the Center for World Indigenous Studies reported that government-sanctioned killings were taking place in the southeastern city of Onitsha, because of a shoot-to-kill policy directed toward Biafrans, particularly members of the MASSOB.[73][74]
The Nigerian federal government accuses MASSOB of violence; MASSOB's leader, Ralph Uwazuruike, was arrested in 2005 and was detained on treason charges. He has since been released and has been rearrested and released more than five times. In 2009, MASSOB leader Chief Uwazuruike launched an unrecognised "Biafran International Passport" and also launched a Biafra Plate Number in 2016 in response to persistent demand by some Biafran sympathizers in the diaspora and at home.[75] On 16 June 2012, a Supr [MAXIMUM LENGTH]

\textbf{AI Message:} unclear
\end{Agentbox}


\end{document}